\documentclass[preprint,12pt,authoryear]{elsarticle}

\usepackage{amssymb}
\usepackage{amsmath}
\usepackage{array}
\usepackage[caption=false,font=normalsize,labelfont=sf,textfont=sf]{subfig}
\usepackage{textcomp}
\usepackage{stfloats}
\usepackage{url}
\usepackage{verbatim}
\usepackage{graphicx}
\usepackage{booktabs}
\usepackage{subfig}
\usepackage{multirow}
\usepackage[linesnumbered,ruled,vlined]{algorithm2e}
\usepackage{natbib}
\usepackage{soul, xcolor}
\usepackage[colorlinks, linkcolor=blue]{hyperref}
\newcommand{\update}[1]{\textcolor{black}{#1}}

\journal{Computer Networks}

\begin{document}

\begin{frontmatter}



\title{Exploiting Features and Logits in Heterogeneous Federated Learning} 


\author{Yun-Hin Chan, and Edith C.H. Ngai} 

\affiliation{organization={Department of Electrical and Electronic Engineering},
            addressline={The University of Hong Kong}, 
            city={Hong Kong},
            country={China}}

\begin{abstract}
Due to the rapid growth of IoT and artificial intelligence, deploying neural networks on IoT devices is becoming increasingly crucial for edge intelligence. Federated learning (FL) facilitates the management of edge devices to collaboratively train a shared model while maintaining training data local and private. However, a general assumption in FL is that all edge devices are trained on the same machine learning model, which may be impractical considering diverse device capabilities. For instance, less capable devices may slow down the updating process because they struggle to handle large models appropriate for ordinary devices. In this paper, we propose a novel data-free FL method that supports heterogeneous client models by managing features and logits, called Felo; and its extension with a conditional VAE deployed in the server, called Velo. Felo averages the mid-level features and logits from the clients at the server based on their class labels to provide the average features and logits, which are utilized for further training the client models. Unlike Felo, the server has a conditional VAE in Velo, which is used for training mid-level features and generating synthetic features according to the labels. The clients optimize their models based on the synthetic features and the average logits. We conduct experiments on two datasets and show satisfactory performances of our methods compared with the state-of-the-art methods. Our codes will be released in the \href{https://github.com/ChanYunHin/Felo\_Velo}{github}.
\end{abstract}


\begin{highlights}
\item We proposed \textbf{Felo}, an FL method for the heterogeneous system environment. It enables clients with different computational resources to select their neural network models with different sizes and architectures. In the Felo, clients exchange their mid-level features and logits based on their class labels to exchange knowledge without a shared public dataset.
\item To fill the knowledge gaps between different client models and extract more latent information from the mid-level features, we proposed \textbf{Velo}, an extension of Felo. The server utilizes conditional VAE to extract more knowledge from the mid-level features.
\item The experiments show that our methods achieve the best performance compared to the state-of-the-art methods. Our methods also outperform FedAvg in the homogeneous environment.
\end{highlights}


\end{frontmatter}



\section{Introduction}
\label{sec:intro}
Machine learning (ML) is playing an increasingly important role in our daily lives. 
It is particularly challenging to deploy ML methods in IoT devices due to their limited computation capabilities, limited network bandwidth, and privacy concerns. Federated learning (FL) \citep{mcmahan2017communication} has been proposed to train neural networks collaboratively in IoT devices (e.g. sensors and mobiles) without communicating private data with each other. 
The first algorithm discussed FL, called FedAvg, coordinates clients and a central server to train a shared neural network but does not require private data to be transmitted to a central server or other clients. In FedAvg, all clients send model weights or gradients to the server after local training, and then the server averages this information to obtain an updated model. This updated model will be sent to the clients, which will continue their training based on the updated model. 

\begin{figure}[!t]
\centering
\includegraphics[width=\textwidth]{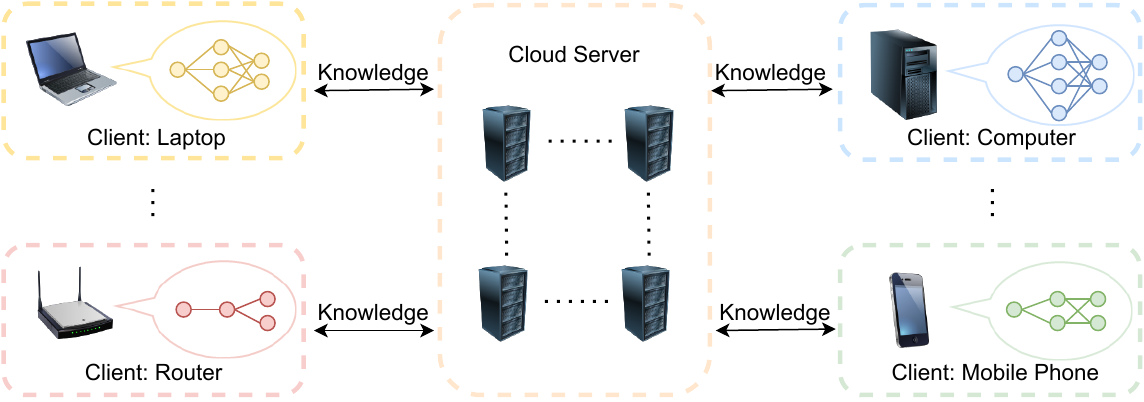}
\caption{\textbf{The problem illustration of system heterogeneity.} These clients are the participants in the federated learning process. The client models of participants are different because of their various available resources. Therefore, the cloud server utilizes shared knowledge from extracted features and logits instead of model weights to update the client models.}
\label{fig_system_heterogeneity}
\end{figure}

However, all the client models have to be identical in FedAvg. The server cannot aggregate and average weights directly if the architectures of the client models are different. Maintaining the same architecture across all models may not be feasible, as it is difficult to assure that all clients have the same computation capabilities, particularly in the IoT environment. System heterogeneity refers to a system containing devices with heterogeneous capabilities as shown in \figurename~\ref{fig_system_heterogeneity}, which is one of the critical challenges in FL for IoT. 
If the clients with different computation capabilities share the same model architecture, the less capable clients may slow down the training speed. 

\begin{figure}[!t]
\centering
\includegraphics[width=\textwidth]{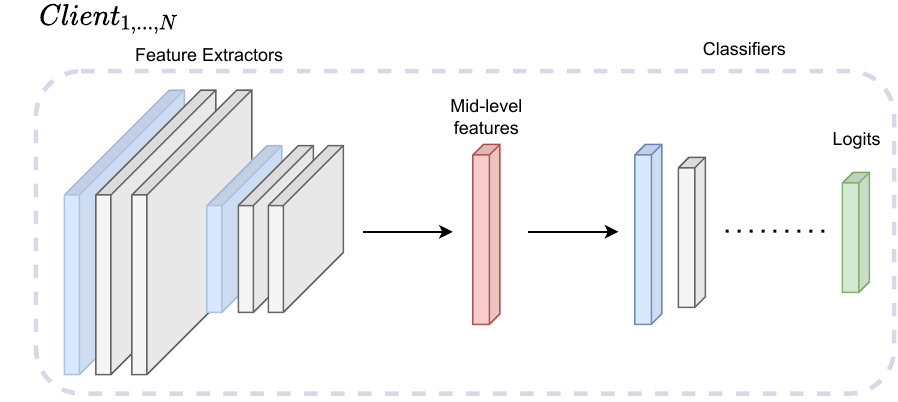}
\caption{The architecture of the client models. A client model is divided into two parts, namely the feature extractor and the classifier. The network architectures of these two parts are not restricted in our design.}
\label{fig_client_models}
\end{figure}

Training heterogeneous models in FL can resolve the system heterogeneity problem as a client can select a model architecture suitable for its computation capability. We summarize recent studies in \tableautorefname~\ref{tab:methods}.
Inspired by knowledge distillation (KD)\citep{hinton2015distilling}, several studies \citep{li2019fedmd}\citep{sattler2021fedaux}\citep{fang2022robust}\citep{huang2022learn} have attempted to manage heterogeneous models in FL. The clients distill knowledge, referred to as logits, from local training, and communicate with each other by logits rather than gradients in these algorithms. \update{Logits are the raw, unnormalized output values produced by the last layer in a neural network, before being passed through the last activation function like softmax, representing the model predictions for each class. \figurename~\ref{fig_client_models} illustrates the position of logits in a neural network. In FL,} 
FedMD \citep{li2019fedmd} incorporates logits derived from a large publicly available dataset. \update{The clients can obtain data features from logits because these logits encode the confidence scores of client models for each class or category, containing rich information about the input data \citep{guo2017calibration}.
MocoSFL \citep{li2023mocosfl} introduces a mechanism that utilizes replay memory on features to enhance KD and MoCo functions \citep{chen2021empirical}, a contrastive framework, in model heterogeneous FL.}
FedAUX\citep{sattler2021fedaux} uses unsupervised pre-training on unlabeled auxiliary data to initialize heterogeneous models in distributed training. RHFL \citep{fang2022robust} deploys the basic knowledge distillation method on the unlabeled public dataset and utilizes the symmetric cross-entropy loss function to compute the weights of different clients in the KD process. 
FCCL \citep{huang2022learn} constructs a cross-correlation matrix on the global unlabeled dataset to exchange information from clients and utilizes KD to alleviate catastrophic forgetting.
\update{HypeMeFed \citep{shin2024effective} introduces a hyper-network and KD approach to generate weights for heterogeneous models in federated learning.}
However, a significant disadvantage of the above methods is that the server has to possess a public dataset, which may not be feasible due to data availability and privacy issues. It can also be difficult for the server to collect sufficient data with the same distribution as the private datasets in the clients. 


Data-free knowledge distillation is a new approach to complete the distillation process without the training data, which is appropriate for FL. The basic idea is to optimize noise inputs to minimize the distance to prior knowledge \citep{nayak2019zero}.
Some studies also tried to utilize data-free knowledge distillation in FL.
FedGen \citep{zhu2021data} uses a generator to simulate the latent features of all the clients. The simulated features are given as prior knowledge of the clients. Then, the client models are updated using their private datasets and these features, though it could be difficult to get a well-trained generator. FedHe \citep{chan2021fedhe} focuses on the logits obtained by the client training process. The server averages these logits, which will be used in the next training round of the clients. FedGKT \citep{he2020group} also does not use a public dataset in the server, but its inference process involves both the clients and the server. The former part of a neural network is conserved in the clients, while the latter is saved in the server. However, this approach drags the speed of a training process and requires additional communication bandwidths during inference.

Except for the methods based on KD, some methods utilize sub-models to handle the system heterogeneity in FL. The clients deploy sub-models from the largest model and train the sub-models based on their local datasets.
HeteroFL \citep{diao2021heterofl} derived different sizes of local models from one large capable neural network model. SlimFL \citep{baek2022joint} incorporated width-adjustable slimmable neural network (SNN) architectures into FL, which can tune the widths of local neural networks. In \citep{horvath2021fjord}, FjORD tailored model widths to clients' capabilities by leveraging Ordered Dropout and a self-distillation methodology. ScaleFL \citep{ilhan2023scalefl} train the splitted client models with cross-entropy and KL-divergence loss. \update{FedASA \citep{deng2024fedasa} uses adaptive model aggregation to handle non-IID data and system heterogeneity. InCoFL \citep{chan2023internal} introduces the gaps in heterogeneous FL, and three splitting methods based on convex optimization to address the gradient divergence problem in heterogeneous FL.}
However, the size of each local model is restricted by the largest neural network model, which means the architecture must be the same except for the number of parameters in each layer.


\begin{table}[!t]
\footnotesize
    \caption{Methods in Heterogeneous Federated Learning}
    \label{tab:methods}
    \centering
    \begin{tabular}{c|cccc}
    \hline
    \hline
        \multirow{2}*{Methods} & \multirow{2}*{Distillation} & Public & \multirow{2}*{Generators} & \multirow{2}*{Sub-models}\\ 
        & & datasets & & \\
      
      \hline
      \multirow{2}*{FedMD \citep{li2019fedmd}}  & \multirow{2}*{\checkmark} & \multirow{2}*{\checkmark} & \multirow{2}*{} & \multirow{2}*{}   \\
         & & &  &    \\
      \hline
      \multirow{2}*{FedAUX \citep{sattler2021fedaux}}  & \multirow{2}*{\checkmark} & \multirow{2}*{\checkmark} & \multirow{2}*{} & \multirow{2}*{}   \\
         & & &  &    \\
      \hline
      \multirow{2}*{FedGen \citep{zhu2021data}}  & \multirow{2}*{\checkmark} & \multirow{2}*{} & \multirow{2}*{\checkmark} & \multirow{2}*{}   \\
         & & &  &    \\
      \hline
      \multirow{2}*{FedHe \citep{chan2021fedhe}}  & \multirow{2}*{\checkmark} & \multirow{2}*{} & \multirow{2}*{} & \multirow{2}*{}   \\
         & & &  &    \\
      \hline
      \multirow{2}*{RHFL \citep{fang2022robust}}  & \multirow{2}*{\checkmark} & \multirow{2}*{\checkmark} & \multirow{2}*{} & \multirow{2}*{}   \\
         & & &  &    \\
      \hline
      \multirow{2}*{FCCL \citep{huang2022learn}}  & \multirow{2}*{\checkmark} & \multirow{2}*{\checkmark} & \multirow{2}*{} & \multirow{2}*{}   \\
         & & &  &    \\
      \hline
      \multirow{2}*{FedGKT \citep{he2020group}}  & \multirow{2}*{\checkmark} & \multirow{2}*{} & \multirow{2}*{} & \multirow{2}*{}   \\
         & & &  &    \\
      \hline
      \multirow{2}*{SlimFL \citep{baek2022joint}}  & \multirow{2}*{} & \multirow{2}*{} & \multirow{2}*{} & \multirow{2}*{\checkmark}   \\
         & & &  &    \\
      \hline
      \multirow{2}*{FjORD \citep{horvath2021fjord}}  & \multirow{2}*{\checkmark} & \multirow{2}*{} & \multirow{2}*{} & \multirow{2}*{\checkmark}   \\
         & & &  &    \\
      \hline
      \multirow{2}*{HeteroFL \citep{diao2021heterofl}}  & \multirow{2}*{} & \multirow{2}*{} & \multirow{2}*{} & \multirow{2}*{\checkmark}   \\
         & & &  &    \\
      \hline
      \multirow{2}*{ScaleFL \citep{ilhan2023scalefl}}  & \multirow{2}*{\checkmark} & \multirow{2}*{} & \multirow{2}*{} & \multirow{2}*{\checkmark}   \\
         & & &  &    \\
      \hline
      \multirow{2}*{\update{MocoSFL \citep{li2023mocosfl}}}  & \multirow{2}*{\update{\checkmark}} & \multirow{2}*{} & \multirow{2}*{} & \multirow{2}*{\update{\checkmark}}   \\
         & & &  &    \\
      \hline
      \multirow{2}*{\update{FedASA \citep{deng2024fedasa}}}  & \multirow{2}*{} & \multirow{2}*{} & \multirow{2}*{} & \multirow{2}*{\update{\checkmark}}   \\
         & & &  &    \\
      \hline
      \multirow{2}*{\update{HypeMeFed \citep{shin2024effective}}}  & \multirow{2}*{\update{\checkmark}} & \multirow{2}*{} & \multirow{2}*{\update{\checkmark}} & \multirow{2}*{\update{\checkmark}}   \\
         & & &  &    \\
      \hline
      \multirow{2}*{\update{InCoFL \citep{chan2023internal}}}  & \multirow{2}*{} & \multirow{2}*{} & \multirow{2}*{} & \multirow{2}*{\update{\checkmark}}   \\
         & & &  &    \\
      \hline
      \multirow{2}*{\textbf{Felo}}  & \multirow{2}*{\checkmark} & \multirow{2}*{} & \multirow{2}*{} & \multirow{2}*{}   \\
         & & &  &    \\
      \hline
      \multirow{2}*{\textbf{Velo}}  & \multirow{2}*{\checkmark} & \multirow{2}*{} & \multirow{2}*{\checkmark} & \multirow{2}*{}   \\
         & & &  &    \\
      \hline
      \hline
    \end{tabular}
\end{table}

To support system heterogeneity and avoid the problems mentioned above, we propose a novel data-free method called \textbf{Felo} and its extension called \textbf{Velo}, which do not require the public dataset or utilize sub-models.
The relations among mid-level features, logits, and the architecture of a client model are shown in \figurename~\ref{fig_client_models}. Felo refers to mid-level \textbf{Fe}atures and \textbf{lo}gits from the client training processes as the exchanged knowledge. At the beginning of Felo, the clients train their models based on their private data and collect the mid-level features and logits from the data according to their class labels. These mid-level features and logits are then transmitted to a server. The server aggregates this information according to their class labels. Finally, this server sends these aggregated features and logits back to clients, which will be utilized to train the client models. The server also aggregates the weights from client models with the same architectures. Throughout the entire process, we assume that the clients have heterogeneous model architectures, which Felo successfully handles the situation. 

However, due to system heterogeneity, the knowledge distillation process is conducted among different model architectures, inducing the knowledge gaps between the large models and small models, which may encumber the KD process \citep{mirzadeh2020improved}. These gaps are filled by our second method, called \textbf{Velo}, which is an extension of Felo. The server deploys a conditional VAE (CVAE) \citep{sohn2015learning} to extract latent relations from the mid-level features in Velo, which can  
enhance the quality of the synthetic features and improve the model accuracy. 
Our experimental results show that Felo and Velo outperform the state-of-the-art algorithms in model accuracy. We further evaluate the accuracy of FedAvg to show that our methods can also achieve satisfactory performance in a homogeneous environment. 
Our contributions are summarized as follows.
\begin{itemize}
    \item We proposed \textbf{Felo}, an FL method for the heterogeneous system environment. It enables clients with different computational resources to select their neural network models with different sizes and architectures. In the Felo, clients exchange their mid-level features and logits based on their class labels to exchange knowledge without a shared public dataset.
    \item To fill the knowledge gaps between different client models and extract more latent information from the mid-level features, we proposed \textbf{Velo}, an extension of Felo. The server utilizes CVAE to extract more knowledge from the mid-level features.
    \item The experiments show that our methods achieve the best performance compared to the state-of-the-art methods. Our methods also outperform FedAvg in the homogeneous environment.
\end{itemize}

\section{Problem Formulation}
We introduce the problem formulation of heterogeneous FL in this section.
In the traditional (homogeneous) federated learning, we have $K$ clients whose neural network architectures are identical. 
Each client has a private dataset $D_k=\{(x_i^k, y_i^k)\}_{i=1}^{|D_k|}$, where $k\in\{1,...,K\}$, and $|D_k|$ means the size of the dataset $D_k$.
Private dataset $D_k$ is only accessible to the client $k$. 
We denote $w$ as the weights of a training model and $f(x;w)$ as a model with weights $w$ and inputs $x$. 
Therefore, the loss function of the client $k$ is $l_k(w)=\frac{1}{|D_k|}\sum_{i=1}^{|D_k|}{l(f(x_i^k;w), y_i^k)}$.
If the total size of all datasets is $N=\sum_{i=1}^K |D_k|$, the global optimization problem is $\sum_{k=1}^K\frac{|D_k|}{N}{l_k(w)}$. 
A common method to solve this problem is to update $w$ according to the aggregation of updated weights and gradients from different clients \citep{mcmahan2017communication}.


The problem of heterogeneous FL is described as follows. One important difference compared to homogeneous FL is that the clients utilize different neural network architectures. We denote $w^k$ as the model weights of the client $k$.
Therefore, the global loss function is $\sum_{k=1}^K\frac{|D_k|}{N}{l_k(w^k)}$ under the heterogeneous model environment. In this case, we need to update weights $\{w^1, w^2, ..., w^K\}$, which are of different sizes, compared to $w$ in the homogeneous FL. Thus, aggregating model weights is impractical with heterogeneous models. Our methods use mid-level features and logits to update $w^k, k\in\{1, ..., K\}$. We will illustrate how to optimize this objective function in the next section.

\section{Our algorithms}
This section presents our data-free methods, Felo and Velo, to solve the heterogeneous FL problem. In heterogeneous FL, model weights can not be aggregated directly due to heterogeneous model architectures at the clients. Hence, the clients exchange knowledge with each other using mid-level features and logits in Felo. 

\figurename~\ref{fig_client_models} shows the architecture of the client models and the positions of the mid-level features and logits. In our methods, the client models are separated into two parts. One is the feature extractor $w_e$, which outputs the mid-level feature from the data input $x_i$. The other is the classifier $w_c$, which predicts the logit of $x_i$ from the mid-level feature. A combination of the feature extractor and the classifier forms a complete neural network. \update{Mid-level features are defined as the outputs of the feature extractor, which processes the input data and generates these features. These features serve as a bridge between the feature extractor and the classifier, which then predicts the logits from the mid-level features.} Note that our algorithms do not require specific model architectures in these two network parts.

\subsection{Felo}
We describe the detailed operations of Felo in this section. The entire algorithm has three steps. 

\subsubsection{Initial training}
At first, clients only train on their private dataset. During the training process, a client $k$ collects mid-level features $s_i^k$, logits $p_i^k$, and the corresponding labels $y_i^k$ from data $(x_i^k, y_i^k)\in D_k$. We regard this collected information as knowledge $(s_i^k, p_i^k, y_i^k)$. After the training process, the client $k$ averages the knowledge by their class labels $y_i^k$, such that the knowledge becomes $(s_{y_i}^k, p_{y_i}^k, y_i^k)$. 

\subsubsection{Average mid-level features and logits}
\begin{figure}[htbp]
\centering
\subfloat[Logits of Felo and Velo]
{\includegraphics[width=\textwidth]{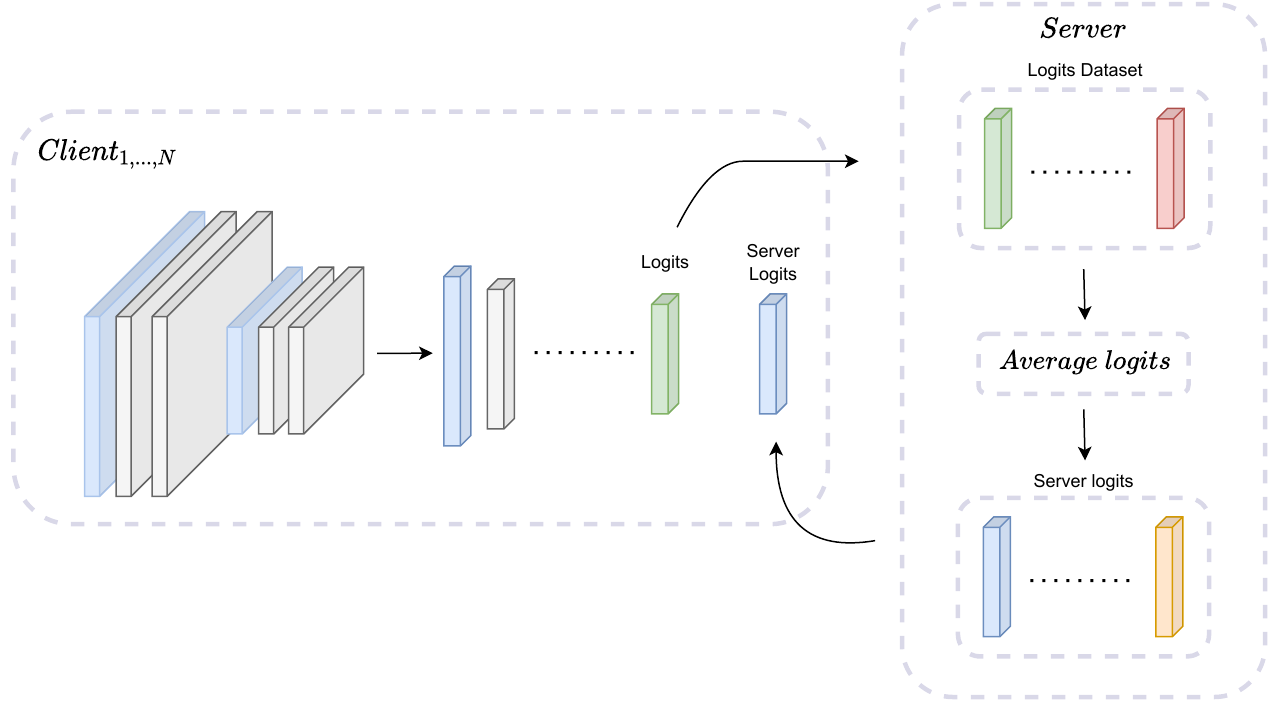}%
\label{fig_logits}}
\hfil
\subfloat[Mid-level features of Velo]
{\includegraphics[width=\textwidth]{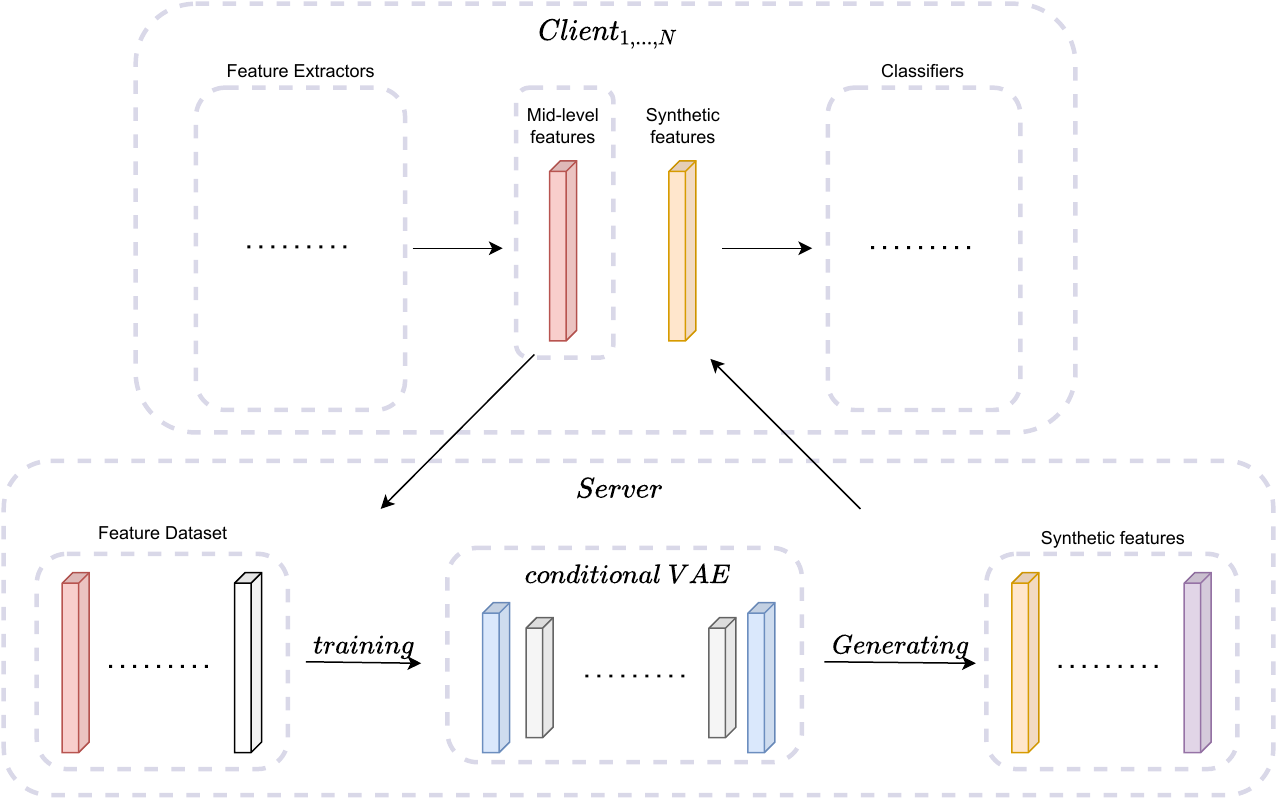}%
\label{fig_velo}}
\caption{The process of handling logits in Felo and Velo is shown in \figurename~\ref{fig_logits}, and the process of handling mid-level features in Velo is shown in \figurename~\ref{fig_velo}. }
\end{figure}

The second step is to average the mid-level features and logits in the server after receiving them from the clients. \figurename~\ref{fig_logits} shows the detailed process for handling the logits. The server averages the logits from the clients based on their classes. It applies the same averaging operation to the features received from the clients. Consider that the server receives knowledge $(s_{y_i}^k, p_{y_i}^k, y_i^k)$ from the client $k$. The knowledge is saved in the server dataset, which is then categorized by the class label. The server averages all the mid-level features and logits received by the clients based on their classes. After averaging, the server obtains $(s_{y_i}^s, p_{y_i}^s, y_i)$, which represents the average features and logits of class $y_i$. The superscript $s$ means these features and logits come from the server. The server will send these per-class average features and logits $(s_{y_i}^s, p_{y_i}^s, y_i)$ back to the clients, which will be used for further updating the client models. The server aggregates and updates the model weights with the same model architectures. The weights from different architectures are not aggregated.

\subsubsection{Training on logits and features}
This step introduces how to train with the logits and mid-level features in the clients. The clients train on their private data and the server's averaged mid-level features and logits during this training process. The client models are divided into two parts, the extractor and the classifier. The averaged mid-level features are the optimized goals for the extractor, while the averaged logits are the optimized objectives for the entire model. The training process is described as follows. The algorithm flow of Felo is shown in Algorithm \ref{Algorithm_Felo}. 

In the beginning, the client $k$ randomly selects a batch of training data $(x_i^k, y_i^k)\in D_k$, and the averaged mid-level features and logits with the same class label are also selected. The training data become $(x_i^k, s_{y_i}^s, p_{y_i}^s, y_i^k)$ after data processing. When the inputs pass through the client model, it can obtain the mid-level features, the logits, and the predicted labels $\Tilde{y_i}=argmax(softmax(p_{y_i}^k))$. Besides the cross-entropy loss function $l_{ce}(\Tilde{y_i}, y_i^k)$, this training process needs two more loss functions to adapt the training for mid-level features and logits. The second loss function, mean square error $l_{mse}(s_{y_i}^k, s_{y_i}^s)$, updates the weights of the extractor. At last, the third loss function, Kullback-Leibler divergence $l_{kl}(p_{y_i}^k, p_{y_i}^s)$, is for the logits. In this case, $l_{kl}$ can update all weights $w^k$ in the client $k$. The loss function of Felo is shown as follows,
\begin{equation}
\label{sum_loss_func_felo}
\begin{aligned}
 L_k=\frac{1}{|\Tilde{D_k}|}\sum_{i=1}^{|\Tilde{D_k}|}{(l_{ce} + \alpha(l_{mse} + l_{kl}})),
\end{aligned}
\end{equation}
where $\alpha$ controls the trade-off between $l_{ce}$ and $l_{mse} + l_{kl}$. This loss function is able to solve the optimization problem of heterogeneous FL. 

\subsection{Velo}
We notice that the server can operate mid-level features and logits when it receives this knowledge from the clients. The server averages the mid-level features and logits but does not explore the latent relationship among mid-level features. 
Moreover, knowledge gaps may occur if the teacher models are vastly larger than the student models \citep{mirzadeh2020improved}, which induced worse learning performances for the student models. This could happen in the FL environment because all the clients would be both teachers and students for each other, which may exacerbate the issue.
Therefore, we deploy a CVAE in the server in \textbf{Velo} to exploit the latent space from the mid-level features. Fig.~\ref{fig_velo} shows the design of the Velo architecture, which is an extension of Felo. Note that the initial training phrase and the training on logits and features in the clients are the same as Felo. The different part is how to train and generate synthetic features on the server. We describe the operations of this part below.

\subsubsection{Training mid-level features with CVAE}
The difference compared to Felo is that CVAE is deployed on the server. Fig.~\ref{fig_velo} describes how to handle mid-level features in Velo. The server still receives knowledge $(s_{y_i}^k, p_{y_i}^k, y_i^k)$ from the clients. Features are collected and saved in the server feature dataset categorized by class labels. Logits are saved in the server logit dataset. The server averages the logits based on their labels, the same as the logit processing in Felo. However, the server uses mid-level features from the feature dataset to train the CVAE.
The training process of the CVAE is similar to \citep{sohn2015learning}. To simplify the notations, we donate the mid-level features as $\textbf{s}$ and their labels as $\textbf{y}$, which are the inputs of the CVAE. The loss function is shown as follows,
\begin{equation}
\label{CVAE_loss_func}
\begin{aligned}
 L_{CVAE}(\textbf{s}, \textbf{y}; \theta, \phi)=-KL(q_\phi(\textbf{z}|\textbf{s},\textbf{y})||p(\textbf{z})) \\
 +\frac{1}{L}\sum_{l=1}^L\log p_\theta(\textbf{s}|\textbf{z}^{(l)},\textbf{y}),
\end{aligned}
\end{equation}
where $\textbf{z}^{(l)}=g_\phi(\textbf{x},\epsilon^{(l)}),\epsilon^{(l)}\sim N(0, \textbf{I}),p(\textbf{z})\sim N(0,\textbf{I})$. The recognition distribution $q_\phi(\textbf{z}|\textbf{s},\textbf{y})$, which is proposed to approximate the true posterior $p_\theta(\textbf{z}|\textbf{s},\textbf{y})$, is reparameterized with an encoder function $g_\phi$, and $p_\theta(\textbf{s}|\textbf{z}^{(l)},\textbf{y})$ is a reconstruction loss of the decoder in CVAE. The first term in Eq. \ref{CVAE_loss_func} aims to project the distribution of mid-level features to $N(0, \textbf{I})$, which is easier for the decoder to generate the synthetic features. The goal of the second term is to recover the original inputs $\textbf{s}$ from $(\textbf{z}^{(l)},\textbf{y})$. 
After completing the training of the CVAE, the server generates synthetic mid-level features for all classes from the decoder $\theta$. These synthetic features and the average logits are sent back to clients. The algorithm flow of Velo is shown in Algorithm \ref{Algorithm_Velo}.


\begin{algorithm}[htbp]
    \caption{Felo}
    \label{Algorithm_Felo}
    \KwIn {Local dataset $D_k, k\in\{1,...,K\}$, $K$ clients, model weights $w_1, ..., w_K$.}
    \KwOut {Optimal weights for all clients $w_1, ..., w_K$.}
    \textbf{Server process:}\\
    \While {not converge}
    {   
        Receive $(s_{y_i}^k, p_{y_i}^k, y_i^k)$ from the client $k$.\\
        Compute the average mid-level features and logits $(s_{y_i}^s, p_{y_i}^s, y_i)$ for each class.\\
        Average the weights from the same model architecture.\\
        Transmit $(s_{y_i}^s, p_{y_i}^s, y_i), \forall y_i$ and the corresponding updated weights to the client requesting for updating.
    }
    \textbf{Client processes:} \\
    \While {random clients $k, k\in {1,...,K}$}
    {
        \If {Initial training} 
        {
            Update the client $k$ model according to $l_{ce}(\Tilde{y_i}, y_i^k)$ from $(x_i^k, y_i^k)\in D_k$.
        }
        \Else
        {
            Receive averaged mid-level features, logits $(s_{y_i}^s, p_{y_i}^s, y_i), \forall y_i$ and the updated weights from the server.\\
            $w_k\leftarrow$ the updated weights.\\
            Update the client $k$ model according to Eq.(\ref{sum_loss_func_felo}).
        }
        Transmit the collected information $(s_{y_i}^k, p_{y_i}^k, y_i^k)$ to the server.
    }
\end{algorithm}

\begin{algorithm}[htbp]
    \caption{Velo}
    \label{Algorithm_Velo}
    \KwIn {Local dataset $D_k, k\in\{1,...,K\}$, $K$ clients and their weights $w_1, ..., w_K$.}
    \KwOut {Optimal weights for all clients $w_1, ..., w_K$.}
    \textbf{Server process:}\\
    \While {not converge}
    {   
        Receive $(s_{y_i}^k, p_{y_i}^k, y_i^k)$ from the client $k$.\\
        Save mid-level features and logits in the server dataset.\\
        Train the CVAE with the inputs $(s_{i}, y_{i})$ from the server dataset according to Eq. (\ref{CVAE_loss_func}). \\
        Average the weights from the same model architecture.\\
        Generate $(s_{y_i}^s, p_{y_i}^s, y_i), \forall y_i$ from the decoder $\theta$. \\
        Send $(s_{y_i}^s, p_{y_i}^s, y_i), \forall y_i$ and the corresponding updated weights to the clients requesting for updating.
    }
    \textbf{Client processes:} \\
    {
        The same as Algorithm\ \ref{Algorithm_Felo}.
    }
\end{algorithm}

\section{Experiments}
In this section, we evaluate the performances of Felo and Velo. We conduct experiments on CIFAR-10 and CINIC-10 datasets under the iid and non-iid settings. CIFAR-10 has 60000 3$\times$32$\times$32 images categorized into ten classes, including 50000 images for training and 10000 images for testing. CINIC-10 is a dataset of the mix of CIFAR-10 and ImageNet, containing a training set, a test set and a validation set, with 90000 images, respectively. Each sample is also a 3$\times$32$\times$32 image with a label from ten classes. We use the training set and the test set in CINIC-10. We use Dirichlet distribution with $\alpha=0.5$ to represent the non-iid data.
For heterogeneous FL, we have five neural network models with different architectures, which are ResNet10, ResNet14, ResNet18, ResNet22 and ResNet26 from the PyTorch source codes. \update{Mid-level features are the outputs of the flatten layer after all CNN blocks. The layers shallower than mid-level features are the feature extractors, and the deeper layers are classifiers, which are defined as one MLP layer in our experiment.} Although our methods support heterogeneous FL, we also conduct experiments on the general homogeneous FL for comparison with FedAvg \citep{mcmahan2017communication}. The model architectures of all clients are ResNet18 in the homogeneous FL experiment. The CVAE deployed in the server of Velo is composed of four dense layers, two for an encoder and two for a decoder. 
Due to the limitation of the memory of our lab server, our experiments include 20 clients for CIFAR-10 and 10 clients for CINIC-10, separated into five groups in FL, i.e. each model architecture is deployed in four clients and two clients, respectively. The client sample ratio is 20\% for all baselines except for FedGKT, which is 100\%, because it cannot converge if the sample ration is 20\% in FedGKT. Moreover, we select the settings of best performances for all baselines under the 20\% sample ratio. 

We implement two baselines \citep{chan2021fedhe} \citep{li2019fedmd} based on the \href{https://fedml.ai/}{FedML} framework. Two baselines \citep{diao2021heterofl} \citep{he2020group} are run on the source codes provided by their authors.

\begin{itemize}
    \item \textbf{FedHe} \citep{chan2021fedhe}: This model focuses on the logits from the clients. The server aggregates and averages the logits and then sends these updated logits back to the clients. The clients train their models based on their private datasets and averaged logits. It is a representative method that only uses KD.
    \item \textbf{FedMD} \citep{li2019fedmd}: This model utilizes logits from a large public dataset for KD. Since it requires a public dataset to complete the training process, the server possesses approximate 10\% of the training dataset to be the public dataset in this baseline. It is a representative method based on utilizing a public dataset and KD.
    \item \textbf{FedGKT} \citep{he2020group}: The clients train small CNN models and periodically send their mid-level features and logits during training processes to the server. The large CNN model in the server continues with the rest of the training process based on these features and logits. It is a representative method with KD and split learning.
    \item \textbf{HeteroFL} \citep{diao2021heterofl}: Four  models with smaller widths but the same depth are sampled from the largest model (ResNet26). These five models are deployed in different clients. This method aggregates the updated parameters from all models to update the largest model. It is a representative method in training with sub-models. The model split mode is dynamic ''a1-b1-c1-d1-e1'' because other algorithms are running five heterogeneous models.
\end{itemize}

It is noteworthy that FedGKT and HeteroFL are evaluated on the server side, which are the performances of ResNet50 and ResNet26. Other methods (including Felo and Velo) evaluate the performance of all clients, which are the averaged performances of different client architectures.
Other hyper-parameters are the default settings in the source code of HeteroFL. 

In the following, we will first compare the model accuracy on CIFAR-10 and CINIC-10 under the iid and non-iid settings. Then, we evaluate the accuracy of FedAvg \citep{mcmahan2017communication} in the homogeneous environment, which shows the powerful performance of Felo and Velo. Experiment results are averaged from three random seeds.

\begin{figure}[!t]
\centering
\subfloat[The average accuracy in iid CIFAR-10]
{\includegraphics[width=0.45\textwidth]{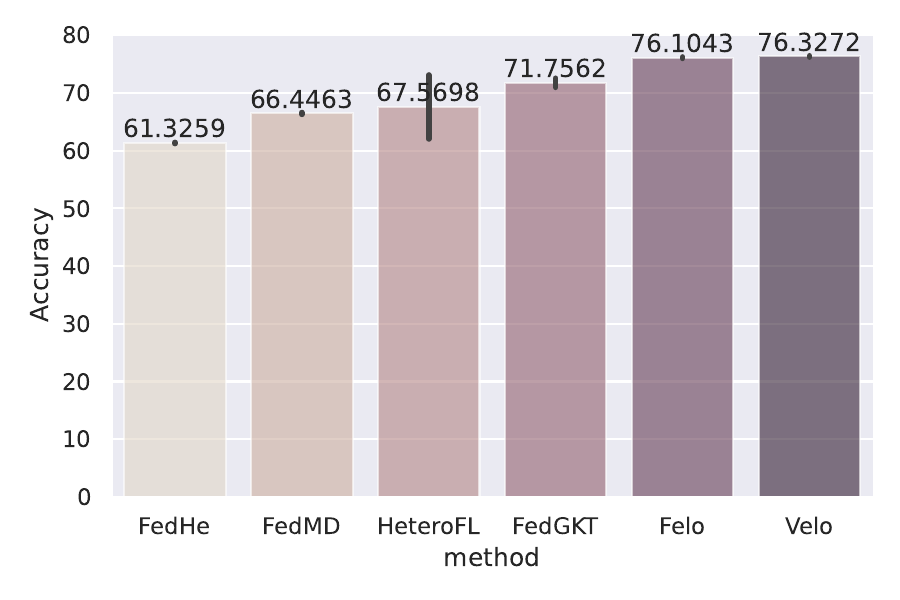}%
\label{fig_bar_iid_cifar10}}
\hfil
\subfloat[The average accuracy in iid CINIC-10]
{\includegraphics[width=0.45\textwidth]{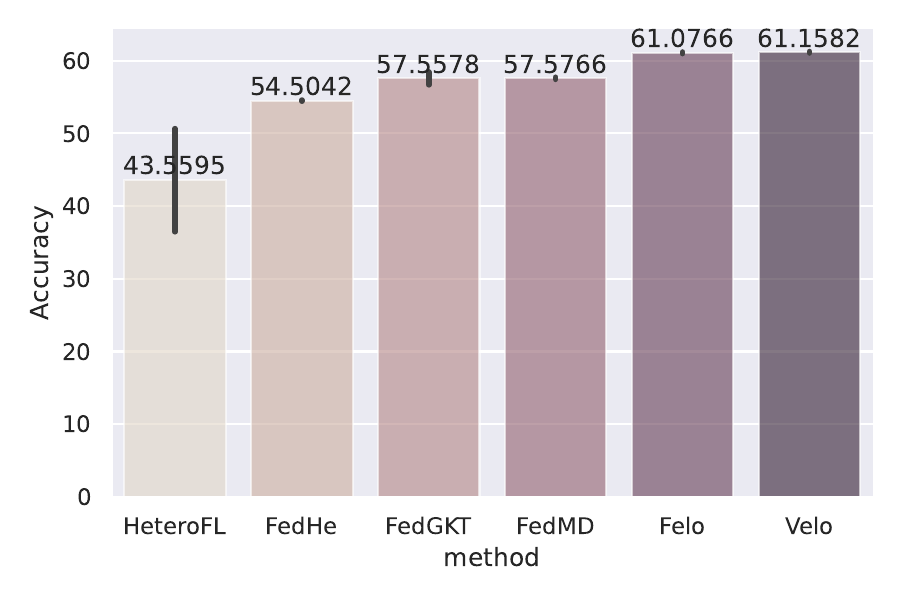}%
\label{fig_bar_iid_cinic10}}

\subfloat[The average accuracy in non-iid CIFAR-10]
{\includegraphics[width=0.45\textwidth]{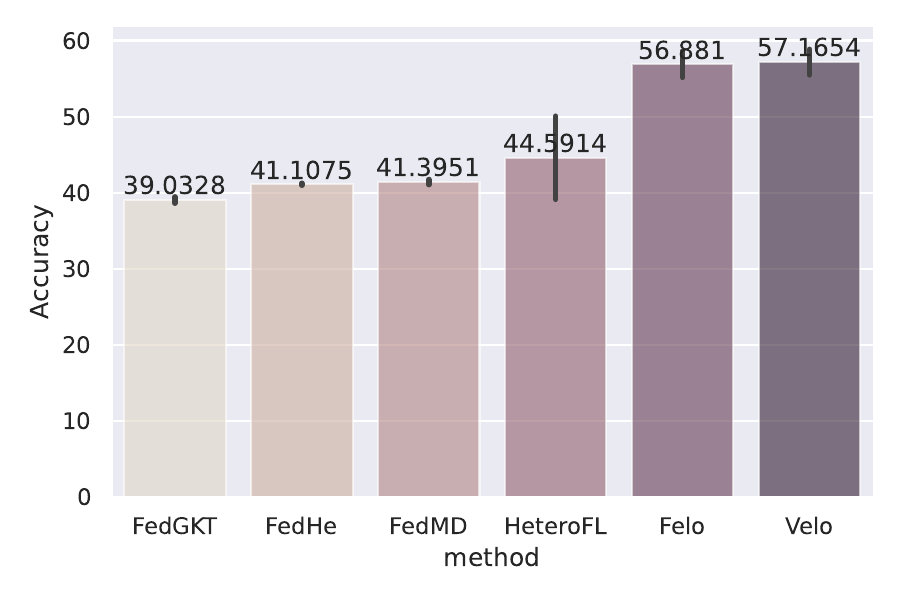}%
\label{fig_bar_noniid_cifar10}}
\hfil
\subfloat[The average accuracy in non-iid CINIC-10]
{\includegraphics[width=0.45\textwidth]{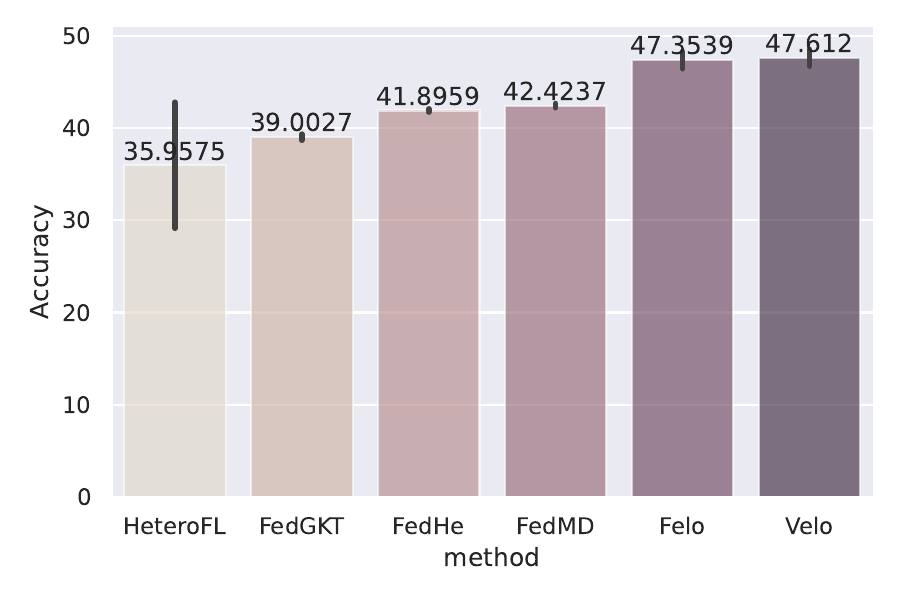}%
\label{fig_bar_noniid_cinic10}}

\caption{Model accuracy of iid and non-iid data in CIFAR-10 and CINIC-10.}
\label{fig_acc}
\end{figure}

\begin{figure}[!t]
\centering
\subfloat[Convergence in iid CIFAR-10]
{\includegraphics[width=0.45\textwidth]{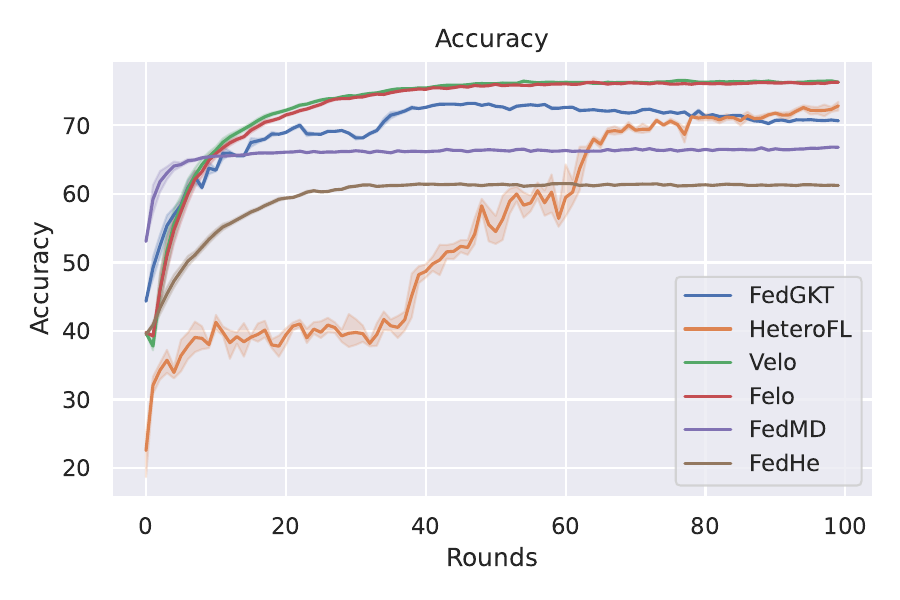}%
\label{fig_iid_cifar10}}
\hfil
\subfloat[Convergence in iid CINIC-10]
{\includegraphics[width=0.45\textwidth]{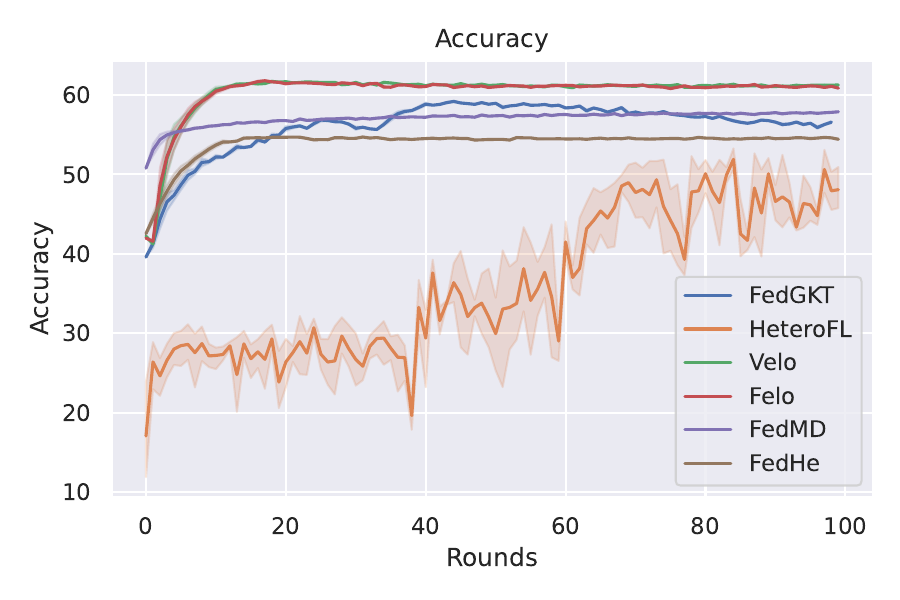}%
\label{fig_iid_cinic10}}

\subfloat[Convergence in non-iid CIFAR-10]
{\includegraphics[width=0.45\textwidth]{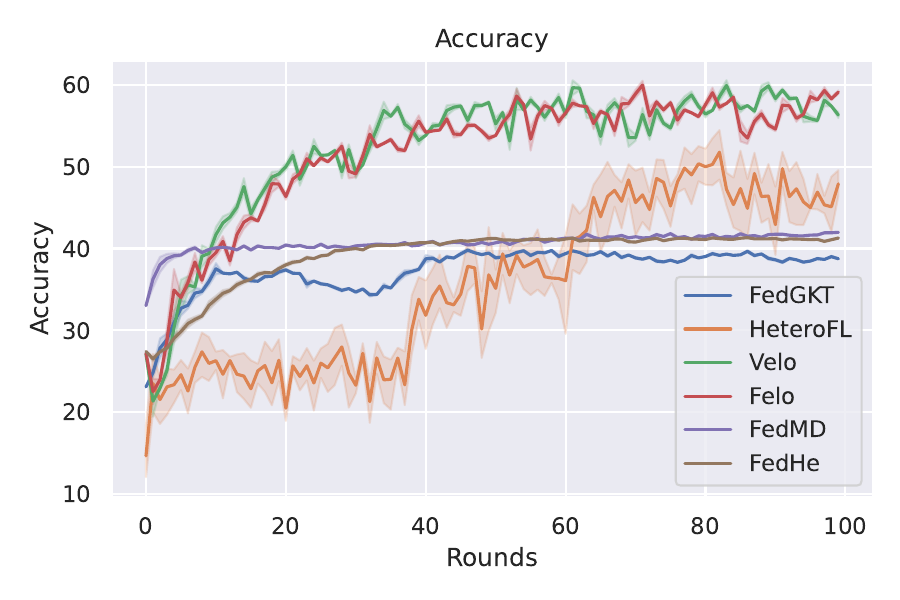}%
\label{fig_noniid_cifar10}}
\hfil
\subfloat[Convergence in non-iid CINIC-10]
{\includegraphics[width=0.45\textwidth]{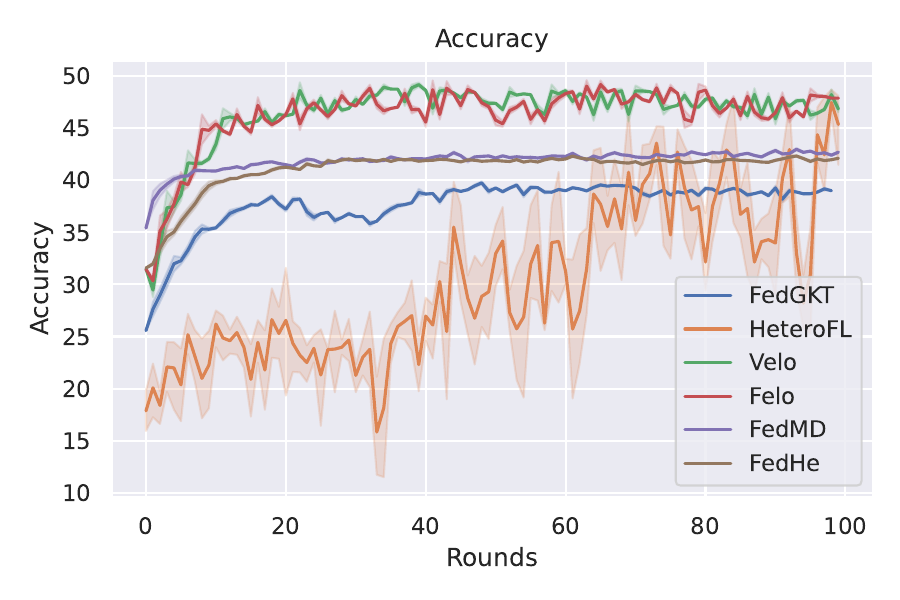}%
\label{fig_noniid_cinic10}}

\caption{Convergence analyses of iid and non-iid data in CIFAR-10 and CINIC-10.}
\label{fig_acc_convergent}
\end{figure}


\subsection{Training on iid datasets}

\figurename~\ref{fig_bar_iid_cifar10} and \figurename~\ref{fig_bar_iid_cinic10} show the average accuracy performance after half of the training process, which is nearly converged. Velo achieves the best performance with an average accuracy of 76.32\% in the iid CIFAR-10 and 61.16\% in the iid CINIC-10. The second-best average accuracy is achieved by Felo, which is 76.10\% in the iid CIFAR-10 and 61.08\% in the iid CINIC-10. 
\figurename~\ref{fig_iid_cifar10} and \figurename~\ref{fig_iid_cinic10} show the training processes of the iid CIFAR-10 and the iid CINIC-10. From \figurename~\ref{fig_iid_cifar10} and \figurename~\ref{fig_iid_cinic10}, the training processes of Felo and Velo are more stable compared to HeteroFL and FedGKT. 
From \figurename~\ref{fig_bar_iid_cifar10} and \figurename~\ref{fig_bar_iid_cinic10}, Velo performs better than Felo, which indicates that Velo can exploit the deeper relations among the mid-level features and reduce the knowledge gaps between large and small models. 
Although HeteroFL and FedGKT are evaluated in the larger models, their performances are worse than Felo and Velo. The reason is that the client models in our methods obtain a better ability than FedGKT to share knowledge by aggregating model weights of the same model architectures. Moreover, our methods consider the entire models leading to more complete knowledge compared to HeteroFL. While FedMD has a public dataset and FedHe utilizes the average logits as shared knowledge, their performances are also worse than our methods. This may be due to Felo utilizes more latent knowledge from the mid-level features and Velo exploits deeper relations among the mid-level features based on the CVAE, compared to FedHe. Although FedMD has the public dataset, this public dataset may not cover all the knowledge from the clients' data. Instead, our methods are able to extract more knowledge from the clients' local data.


\subsection{Training on non-iid datasets}
Apart from iid datasets, we conduct non-iid experiments on CIFAR-10 and CINIC-10 as shown in \figurename~\ref{fig_bar_noniid_cifar10},  \figurename~\ref{fig_bar_noniid_cinic10}, \figurename~\ref{fig_noniid_cifar10} and \figurename~\ref{fig_noniid_cinic10}. We notice that the training processes of Felo and Velo are unstable compared to the performance in the iid dataset, indicating the knowledge from mid-level features are more biased. However, the results of our methods are still the best among these state-of-the-art algorithms. 
The average accuracy in non-iid settings are shown in \figurename~\ref{fig_bar_noniid_cifar10} and \figurename~\ref{fig_bar_noniid_cinic10}.
According to these two figures, the best results still come from Velo, 57.17\% in non-iid CIFAR-10 and 47.61\% in non-iid CINIC-10. The second one is 56.88\% in non-iid CIFAR-10 and 47.35\% in  non-iid CINIC-10 from Felo.
It is worth mentioning that the performances of FedHe, FedMD and FedGKT in non-iid CIFAR-10 are worse than their performances in non-iid CINIC-10, since the client number is 20 in  CIFAR-10 while it is 10 in CINIC-10. Increasing the client number implies more extreme distributions among clients, which are more difficult to be handled by these algorithms. Nevertheless, our methods still obtain the best performances compared with other methods when handling these challenging environments.

In non-iid settings, the data distribution becomes more complex and difficult to handle, but our methods also perform better. These results prove that Felo can share better knowledge than other methods. Furthermore, Velo is able to exploit the latent information among the mid-level features to improve the quality of synthetic mid-level features, which also reduces the knowledge gaps among different model architectures, leading to better client models and better performances.

\begin{table*}[!t]
\footnotesize
    \caption{Accuracy of FedAvg, Felo, and Velo with homogeneous models.}
    \label{tab:homo_acc}
    \centering
    \begin{tabular}{c|cccc}
    \toprule
         \multirow{2}*{Method} &\multicolumn{2}{c}{CIFAR-10} & \multicolumn{2}{c}{CINIC-10} \\
      \cline{2-5}
       & iid & non-iid & iid & non-iid   \\
      \midrule
       \multirow{2}*{FedAvg} & \multirow{2}*{84.582$\pm$0.26\%} & \multirow{2}*{59.229$\pm$0.22\%} & \multirow{2}*{70.26$\pm$0.09\%} & \multirow{2}*{46.837$\pm$0.45\%}   \\
          &  &  &  &  \\
      \hline
       \multirow{2}*{Felo} & \multirow{2}*{84.451$\pm$0.41\%} & \multirow{2}*{60.357$\pm$0.52\%} & \multirow{2}*{70.697$\pm$0.12\%} & \multirow{2}*{\textbf{48.528$\pm$0.40\%}}   \\
          &  &  &  &  \\
      \hline
       \multirow{2}*{Velo} & \multirow{2}*{\textbf{85.077$\pm$0.32\%}} & \multirow{2}*{\textbf{60.882$\pm$0.40\%}} & \multirow{2}*{\textbf{70.858$\pm$0.15\%}} & \multirow{2}*{47.885$\pm$0.48\%}   \\
          &  &  &  &  \\
    \bottomrule
    \end{tabular}
\end{table*}

\subsection{Compared to FedAvg}
We also conduct homogeneous FL experiment to compare with FedAvg. 
We use ResNet18 in this experiment. To show the results in a more stable environment, we set the sample ratio to 100\% in the non-iid environment. The results are shown in \tableautorefname~\ref{tab:homo_acc}. No matter using iid or non-iid data in the two datasets, our methods obtain better performances than FedAvg. Velo obtains the best performance, 85.07\%, which is better than Felo (84.45\%) and FedAvg (84.58\%) in iid CIFAR-10. For non-iid CIFAR-10, the best performance is still from Velo, which is 60.88\%. The second-best is 60.36\% from Felo, and the last one is 59.23\% from FedAvg. In the iid CINIC-10 dataset, Velo achieves the best accuracy which is 70.86\%, and the following are 70.70\% from Felo and 70.26\% from FedAvg. Felo achieves the best result, 48.53\%, in  non-iid CINIC-10. The second one is 47.89\% from Velo, and the last one is 46.84\% from FedAvg. Our methods perform better than FedAvg in different environments, proving the improvements from exploiting mid-level features and logits.
Furthermore, the additional communication overheads for exchanging averaged mid-level features and logits in our methods are negligible, which are smaller than 0.05\% of the overheads in FedAvg. 


\subsection{\update{Experiments on extreme heterogeneous models}}
\update{The goal of Velo is to exploit the latent space from the mid-level features. To demonstrate the effect of cVAE in Velo, we conduct experiments on more extreme heterogeneous models, i.e., five different model architectures, which are ResNet10, ResNet18, ResNet26, ResNet34, and ResNet50. Our previous model architectures are ResNet10, ResNet14, ResNet18, ResNet22, and ResNet26.}

\update{Table~\ref{tab:extreme_heter_acc} shows the experiment results for extreme heterogeneous models. Felo and Velo still achieve better performance than other baselines, such as FedHe and FedGKT, across both CIFAR-10 and CINIC-10 datasets under both iid and non-iid conditions.
For instance, Velo (Extreme) achieves the highest accuracy among the extreme heterogeneous models, with 74.13\% on iid CIFAR-10 and 60.24\% on iid CINIC-10.
Moreover, the performance degradation of Velo is smaller than Felo, demonstrating that cVAE on the server exploits the latent relationship between mid-level features and generates synthetic features to fill the knowledge gaps.
These results suggest that Velo is more robust in handling model heterogeneity, further validating the effectiveness of leveraging cVAE for feature extraction.
}

\begin{table*}[!t]
\footnotesize
    \caption{\update{Accuracy of extreme heterogeneous models. \textit{Previous} refers to the model architectures in previous experiments, and \textit{Extreme} denotes the extreme heterogeneous model architectures for this experiment.}}
    \label{tab:extreme_heter_acc}
    \centering
    \begin{tabular}{ccccc}
    \toprule
         \multirow{2}*{Method} &\multicolumn{2}{c}{CIFAR-10} & \multicolumn{2}{c}{CINIC-10} \\
      \cline{2-5}
       & iid & non-iid & iid & non-iid   \\
      \midrule
       \multirow{2}*{FedHe} & \multirow{2}*{55.16$\pm$1.87\%} & \multirow{2}*{38.97$\pm$2.34\%} & \multirow{2}*{50.25$\pm$1.48\%} & \multirow{2}*{37.81$\pm$1.78\%}   \\
          &  &  &  &  \\
      \multirow{2}*{FedGKT} & \multirow{2}*{65.82$\pm$1.34\%} & \multirow{2}*{37.13$\pm$1.56\%} & \multirow{2}*{55.30$\pm$1.49\%} & \multirow{2}*{35.62$\pm$1.73\%}   \\
          &  &  &  &  \\
       \multirow{2}*{Felo (Extreme)} & \multirow{2}*{70.451$\pm$1.27\%} & \multirow{2}*{52.63$\pm$1.86\%} & \multirow{2}*{57.27$\pm$1.22\%} & \multirow{2}*{44.23$\pm$1.49\%}   \\
          &  &  &  &  \\
       \multirow{2}*{Velo (Extreme)} & \multirow{2}*{\textbf{74.13$\pm$1.51\%}} & \multirow{2}*{\textbf{56.92$\pm$1.89\%}} & \multirow{2}*{\textbf{60.24$\pm$1.14\%}} & \multirow{2}*{\textbf{46.03$\pm$1.77\%}}   \\
          &  &  &  &  \\
       \hline
       \multirow{2}*{Felo (Previous)} & \multirow{2}*{76.10$\pm$1.01\%} & \multirow{2}*{56.88$\pm$1.52\%} & \multirow{2}*{61.08$\pm$0.58\%} & \multirow{2}*{47.35$\pm$1.30\%}   \\
          &  &  &  &  \\
       \multirow{2}*{Velo (Previous)} & \multirow{2}*{76.33$\pm$1.24\%} & \multirow{2}*{57.17$\pm$1.40\%} & \multirow{2}*{61.16$\pm$0.96\%} & \multirow{2}*{47.61$\pm$1.33\%}   \\
          &  &  &  &  \\
    \bottomrule
    \end{tabular}
\end{table*}

\subsection{\update{Experiments on cVAE training interval.}}
\update{In Velo, the cVAE is trained on the server, a process that requires computational time. This training phase, however, has the potential to introduce delays into the entire FL process. To address this concern, we conduct experiments to evaluate the impact of varying training intervals for the cVAE on the system performance, with the aim of mitigating delays caused by its training. Specifically, we hypothesize that in scenarios where the server is constrained by limited computational resources, increasing the interval between cVAE training iterations, rather than performing training at every communication round, could effectively reduce delays.}

\update{Table~\ref{tab:cvae_interval} presents the experimental results for different training intervals. Notably, across intervals ranging from 1 to 10 communication rounds, Velo demonstrates stable performance on both IID and Non-IID CIFAR-10 datasets. This stability suggests that Velo is robust to extended cVAE training intervals, maintaining its effectiveness even when the cVAE is updated less frequently. This finding implies that in resource-constrained environments, the server can strategically increase the cVAE training interval to mitigate delays without significantly compromising model performance. Furthermore, the server can leverage parallelization by training the cVAE concurrently with the local training of client models, thereby enhancing the overall efficiency of the training process. This parallel training mechanism not only reduces idle time but also optimizes resource utilization, ensuring that the cVAE training does not become a bottleneck in the FL.}

\begin{table*}[!t]
\footnotesize
    \caption{\update{Accuracy of different training intervals for cVAE. The value of total communication rounds is 100.}}
    \label{tab:cvae_interval}
    \centering
    \begin{tabular}{ccccccccc}
    \toprule
         \multirow{2}*{CIFAR-10} &\multicolumn{7}{c}{CVAE training interval in Velo (rounds)} \\
      \cline{2-9}
       & 1 & 2 & 3 & 4 & 5 & 6 & 8 & 10  \\
      \midrule
       \multirow{2}*{IID} & \multirow{2}*{76.33} & \multirow{2}*{75.52} & \multirow{2}*{76.81} & \multirow{2}*{75.79} & \multirow{2}*{76.40} & \multirow{2}*{75.89} & \multirow{2}*{77.37} & \multirow{2}*{75.91}  \\
          &  &  &  &  \\
       \multirow{2}*{Non-IID} & \multirow{2}*{57.17} & \multirow{2}*{58.02} & \multirow{2}*{58.72} & \multirow{2}*{56.53} & \multirow{2}*{58.39} & \multirow{2}*{57.46} & \multirow{2}*{55.24} & \multirow{2}*{57.75}    \\
          &  &  &  &  \\
    \bottomrule
    \end{tabular}
\end{table*}

\section{Conclusions}
In conclusion, we proposed a novel data-free heterogeneous FL method called Felo and its extension called Velo. Our methods support FL training on heterogeneous models, which are practical for client devices with heterogeneous computation capabilities. Other than model weights, logits and mid-level features are collected by the client training processes and sent to the server. The server averages the logits and  mid-level features based on their class labels in Felo. For the extension in Velo, the server further utilizes the mid-level features from the clients to train a conditional VAE to generate synthetic features. The clients then perform local training with their private datasets, together with the server's averaged logits and synthetic features. We conducted experiments on heterogeneous FL with two datasets, which demonstrate the high accuracy of our methods compared with the state-of-the-art. We further compared our methods with FedAvg in homogeneous environment, proving the superior performance of our methods.

In the future, we will reform the interactions between the clients and the server in Velo to facilitate parallel training for the clients and the server. The clients can obtain new synthetic features and logits continually to train their models. Meanwhile, the server continues with its training of the CVAE after adding the new mid-level features from the clients. This parallel implementation can increase the training epochs of the CVAE on the server and advance the quality of the synthetic features.









\bibliographystyle{elsarticle-harv} 
\bibliography{ref}

\end{document}